\documentclass[runningheads]{llncs}

\usepackage[T1]{fontenc}
\usepackage{amsmath,amssymb,amsfonts}
\usepackage{booktabs}
\usepackage{multirow}
\usepackage{graphicx}
\usepackage{float}
\usepackage{subcaption}
\usepackage{hyperref}
\usepackage{xcolor}
\usepackage{enumitem}

\begin{document}

\title{Learning Coarse-to-Fine Osteoarthritis Representations under Noisy Hierarchical Labels}
\titlerunning{Learning Osteoarthritis Representations under Hierarchical Labels}

\author{Tongxu Zhang}
\authorrunning{T. Zhang}
\institute{The Hong Kong Polytechnic University \\
\email{jukie.zhang@connect.polyu.hk}}

\maketitle

\begin{abstract}
Knee osteoarthritis (OA) assessment contains a natural label hierarchy between binary disease status and Kellgren--Lawrence (KL) severity. We study whether supervision at these two granularities changes learned 3D MRI representations. A shared encoder with OA and KL prediction heads is evaluated under single-OA, single-KL, and dual-head training across ResNet3D, M3T, and nnMamba backbones. Evaluation combines predictive metrics with paired statistical comparisons under Benjamini--Hochberg FDR control, latent severity-axis geometry, and saliency--cartilage overlap. Dual supervision yields significant KL-grading gains for ResNet3D and a significant OA AUC gain for M3T, whereas nnMamba retains stronger single-task performance. Representation analysis further shows an architecture-dependent effect: Dual strengthens label-aligned latent organization for ResNet3D and M3T, while nnMamba retains stronger alignment under single-task supervision. For the responsive backbones, Dual also produces descriptively higher saliency overlap with cartilage. These results show that coarse-to-fine supervision can reshape disease representations under noisy hierarchical labels, with benefits that depend on backbone architecture. Code is available at \url{https://github.com/jukieCheung/coarse2fine-oa-mri}.

\end{abstract}

\section{Introduction}

Knee osteoarthritis (OA) assessment contains a clinically meaningful coarse-to-fine label hierarchy. A binary OA decision describes disease presence, while the Kellgren--Lawrence (KL) grade provides a finer ordinal description of structural severity \cite{kellgren1957radiological}. These targets are closely related but impose supervision at different granularities. KL grading is semi-quantitative and reader-dependent, with inter- and intra-observer variability concentrated around adjacent grades \cite{kohn2016classifications,kose2018reliability}. Binary OA supervision provides a coarser and more stable disease partition, while KL supervision preserves fine-grained severity information.

Most deep-learning studies evaluate OA classification and KL grading primarily through scalar predictive metrics. This leaves an important representation-learning question insufficiently explored: how does supervision at different clinical granularities organize disease information in the latent space? A model optimized only for binary OA discrimination may emphasize a strong disease-presence boundary and compress within-class severity variation. Fine-grained KL supervision encourages separation among five severity categories but must learn directly from noisy adjacent-grade boundaries. Joint supervision therefore provides a controlled setting for examining whether a coarse disease objective can stabilize the representation while preserving finer severity-related structure.

The relationship between the two targets deserves particular attention because the binary OA label in this study is derived from the KL grade. The two tasks do not introduce independent annotation information. Their potential value arises from the optimization constraints they impose on the shared encoder: KL supervision separates five severity categories, whereas OA supervision explicitly aggregates grades 0--1 and 2--4 into a clinically meaningful coarse partition. We therefore study whether adding this coarse-granularity objective changes predictive behavior and latent organization under the same underlying label hierarchy.

Beyond latent organization, a clinically useful representation should also exhibit plausible spatial attribution. OA involves multiple structures, including cartilage, subchondral bone, osteophytes, menisci, and joint-space-related changes, while cartilage remains a central tissue in structural OA assessment \cite{hunter2011evolution,roemer2014anterior}. We therefore complement predictive and latent-space analyses with saliency--cartilage overlap as a focused post-hoc measure of anatomical plausibility.

\begin{figure}[t]
    \centering
    \includegraphics[width=0.93\textwidth]{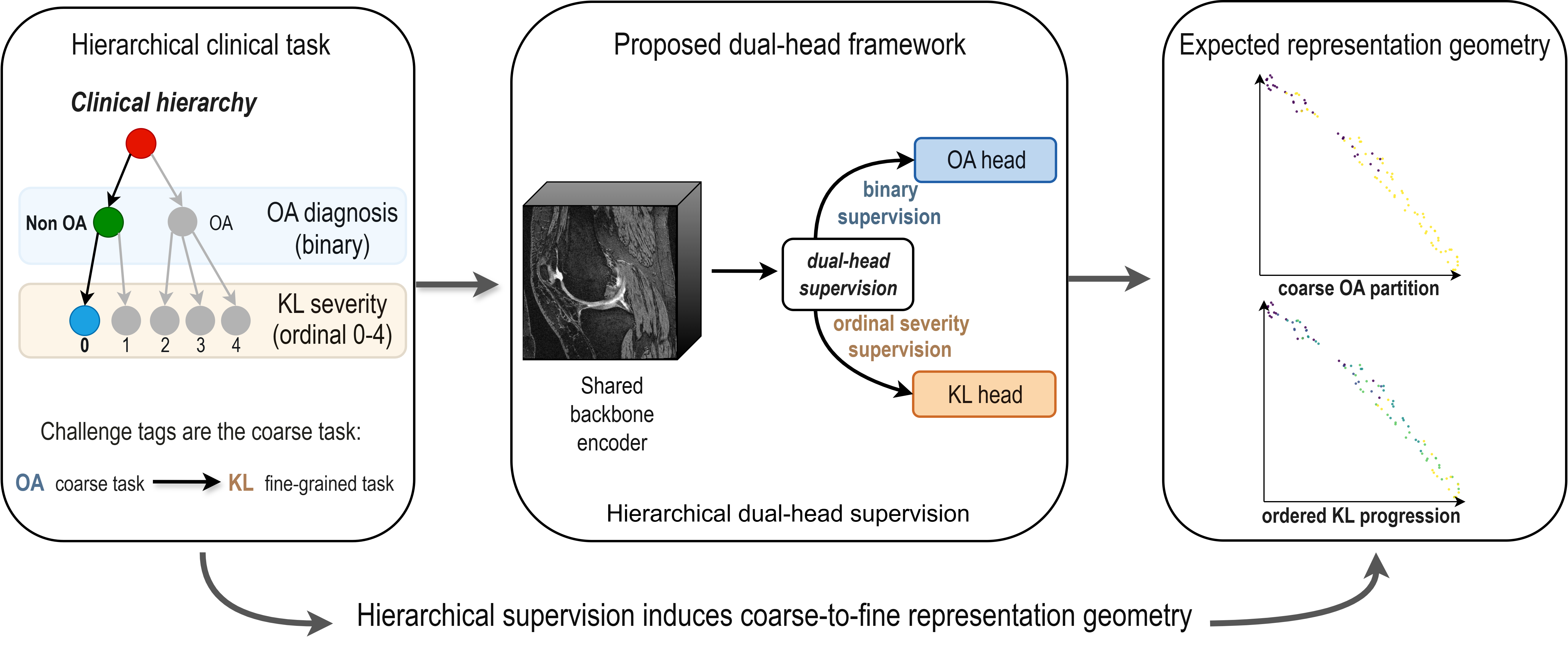}
    \caption{Study overview. Binary OA diagnosis and ordinal KL grading define two supervisory granularities within the same clinical hierarchy. Single-task and dual-head training are compared across three 3D backbones using predictive, representation-level, and anatomical attribution analyses.}
    \label{fig:framework}
\end{figure}

The study makes three contributions. First, we formulate OA/KL learning as a coarse-to-fine representation problem and explicitly examine the effect of supervision granularity when the coarse label is derived from the fine label. Second, we compare Single-OA, Single-KL, and Dual training across CNN-, Transformer-, and state-space-based 3D backbones under an identical subject-level protocol with paired statistical testing. Third, we connect predictive effects to latent severity-axis geometry and saliency--cartilage overlap, enabling a representation-level analysis of when hierarchical supervision is beneficial and when it is not.

\section{Related Work}

\subsection{Deep learning for knee OA assessment}

Deep learning has been widely applied to knee OA assessment from radiographs and MRI, including binary OA classification, KL grading, structural severity assessment, and progression prediction \cite{vaattovaara2025kl,panwar2024mri,alyami2024mri}. Most studies formulate disease presence and severity grading as predictive tasks and evaluate performance using accuracy, AUC, or F1 score. Multi-task learning offers a general framework for coupling such related objectives through a shared representation \cite{caruana1997multitask,ruder2017overview}. These studies establish the value of shared feature learning, while the effect of supervision design on the organization of the learned disease representation remains less directly characterized.

\subsection{Ordinal KL labels and hierarchical supervision}

KL grading has an ordered category structure and is subject to reader variability, particularly between adjacent grades \cite{kohn2016classifications,kose2018reliability}. Ordinal-aware formulations can explicitly encode class order, while hierarchical classification studies emphasize semantic relationships between coarse and fine categories \cite{huo2019coarse,park2024hierarchical,lang2024coarse}. In knee OA, binary OA status and KL severity form a natural two-level hierarchy: grades 0--1 belong to the OA-negative branch and grades 2--4 to the OA-positive branch. This setting allows the effect of supervisory granularity to be examined without changing the underlying clinical annotation source.

Multi-task learning provides a natural mechanism for coupling related objectives through a shared representation \cite{caruana1997multitask,ruder2017overview}. The present study uses the dual-head setting as a controlled supervision design: the OA objective emphasizes the coarse disease boundary and the KL objective retains fine-grained severity distinctions. The central empirical question is whether these two granularities reshape the shared feature space differently across backbone architectures.

\subsection{Representation geometry and anatomical plausibility}

Representation geometry provides information that is not captured by predictive metrics alone. Analyses of latent axes and label alignment can reveal whether disease severity is organized coherently in feature space \cite{bengio2013representation,li2022neuralmanifold}. For hierarchical OA assessment, a useful representation should preserve both coarse OA separability and fine-grained severity structure.

Post-hoc attribution provides a complementary spatial view of model behavior. Saliency and class-activation methods are commonly used to inspect whether predictions are associated with relevant image regions \cite{simonyan2013deep,selvaraju2017grad,smilkov2017smoothgrad}. We use cartilage-focused overlap as one anatomically motivated probe of attribution patterns. Because cartilage represents only one component of OA pathology and input-gradient attribution has known limitations, this analysis is interpreted as descriptive supporting evidence with a deliberately limited explanatory scope.

\section{Method}

\subsection{Problem formulation}

Let $x$ denote a knee MRI volume, $y^{\mathrm{KL}}\in\{0,1,2,3,4\}$ the KL grade, and $y^{\mathrm{OA}}\in\{0,1\}$ the binary OA label. The coarse label is derived from the KL hierarchy as
\begin{equation}
    y^{\mathrm{OA}}=\mathbb{I}(y^{\mathrm{KL}}\ge2),
\end{equation}
so grades $\{0,1\}$ form the OA-negative branch and grades $\{2,3,4\}$ form the OA-positive branch.

We compare three supervision settings: \textbf{Single-OA}, with one OA classification head; \textbf{Single-KL}, with one KL grading head; and \textbf{Dual}, with both heads sharing the same encoder. For a latent representation $z=f_{\theta}(x)$, the two heads predict
\begin{equation}
    \hat{y}^{\mathrm{OA}}=g_{\phi}(z),\qquad
    \hat{y}^{\mathrm{KL}}=h_{\psi}(z).
\end{equation}
The Dual model minimizes
\begin{equation}
    \mathcal{L}=\lambda_{\mathrm{OA}}\mathcal{L}_{\mathrm{OA}}
    +\lambda_{\mathrm{KL}}\mathcal{L}_{\mathrm{KL}},
\end{equation}
where $\mathcal{L}_{\mathrm{OA}}$ is binary cross-entropy with logits and $\mathcal{L}_{\mathrm{KL}}$ is multi-class cross-entropy.

Because $y^{\mathrm{OA}}$ is deterministically derived from $y^{\mathrm{KL}}$, the two objectives use the same annotation hierarchy. Their distinction lies in supervision granularity. The KL objective separates five severity categories, whereas the OA objective explicitly aggregates grades 0--1 and 2--4 into a coarse clinical partition. This changes the loss geometry seen by the shared encoder and may reinforce the clinically meaningful OA boundary while allowing the KL head to retain within-branch severity variation. The experiments test whether this additional coarse-granularity constraint changes representation learning across different 3D architectures.

The KL head also implies a coarse OA probability,
\begin{equation}
    p^{\mathrm{OA}}_{\mathrm{from\,KL}}=\sum_{k=2}^{4}p^{\mathrm{KL}}_k,
\end{equation}
which makes the hierarchical relationship explicit at the prediction level. The current formulation uses the two task losses without an explicit consistency penalty, allowing the empirical comparison to focus on the effect of adding the coarse objective itself.

\subsection{Dataset and experimental protocol}

Experiments use OAIZIB-CM \cite{peterfy2008osteoarthritis,ambellan2019automated,yao2024cartimorph} with predefined subject-level splits of 383 training and 98 held-out test subjects. Each subject contributes one examination entry. Training KL counts for grades 0--4 are 82/46/86/111/58 and test counts are 21/12/22/28/15. Applying the OA threshold yields 128 OA-negative and 255 OA-positive training cases, and 33 OA-negative and 65 OA-positive test cases. The same split is used for every supervision setting and backbone.

The three backbone families are a 3D ResNet-18 \cite{he2016deep}, the M3T medical-image Transformer \cite{jang2022m3t}, and nnMamba \cite{gong2024nnmamba}. These architectures provide CNN-, Transformer-, and state-space-based instantiations of the same supervision design. The 3D ResNet-18 and nnMamba are trained from random initialization; the 2D multi-plane branch of M3T is initialized with ImageNet-pretrained ResNet-18 weights \cite{he2016deep}, following its published implementation. Each single-task model uses the same backbone followed by one task-specific head, while Dual uses two parallel heads on the shared encoder.

MRI volumes are loaded from NIfTI files, normalized using per-volume z-score normalization, and rearranged to $[D,H,W]$. All volumes are sagittal DESS acquisitions as documented for the OAI-ZIB dataset \cite{ambellan2019automated}. ResNet3D and nnMamba use $160\times256\times256$ inputs, while M3T uses $128^3$ inputs to match its implementation constraints. All models are trained for 100 epochs with batch size 2, AdamW, learning rate $10^{-4}$, and weight decay $10^{-4}$. The Dual objective uses equal loss weights ($\lambda_{\mathrm{OA}}=\lambda_{\mathrm{KL}}=1$). No learning-rate scheduler, data augmentation, class weighting, label smoothing, or early stopping is used; the final-epoch checkpoint is evaluated, and no model selection is performed on the held-out test set.

\subsection{Evaluation metrics}

For OA classification, we report AUC, accuracy, and F1 score. For KL grading, we report macro-averaged one-vs-rest AUC, accuracy, and macro-F1. Macro averaging reduces dominance by larger KL classes.

For representation geometry, penultimate-layer features are extracted for all training and test subjects and analyzed with PCA. Feature standardization and PCA are fitted on the 383 training subjects only, and the fitted transform is applied to the 98 held-out test subjects. We report the explained variance ratio of PC1 and the absolute Spearman correlation magnitudes $|\rho(\mathrm{PC1},\mathrm{KL})|$ and $|\rho(\mathrm{PC1},\mathrm{OA})|$ computed on the test subjects. The absolute value is used because the sign of a PCA eigenvector is arbitrary; the metric therefore captures the strength of monotonic label alignment without assigning semantic meaning to axis orientation. Coarse separability is summarized by a one-dimensional logistic-regression probe on the PC1 score, fitted on the training subjects and evaluated as AUROC on the held-out test subjects. No supervised model is fitted on test features.

For anatomical attribution, saliency maps are computed with SmoothGrad \cite{smilkov2017smoothgrad}: the gradient of the target logit with respect to the input MRI is averaged over 8 Gaussian-noise-perturbed copies (noise standard deviation 0.05), multiplied element-wise by the input (gradient$\times$input), followed by absolute value and per-volume min-max normalization to $[0,1]$. For OA, the positive-class logit is used; for KL, the predicted-class logit is used. The cartilage ROI $M$ is the union of the femoral and tibial cartilage labels (segmentation labels 2 and 4) of the OAI-ZIB segmentations \cite{ambellan2019automated}. Let $S_i$ denote voxel-wise saliency, $M_i$ the binary cartilage mask, and $T_{k\%}$ the indicator of the top $k\%$ salient voxels within each volume. We compute
\begin{equation}
\mathrm{mass@ROI}=\frac{\sum_i S_iM_i}{\sum_i S_i},\qquad
\mathrm{top1@ROI}=\frac{\sum_iT_{1\%,i}M_i}{\sum_iT_{1\%,i}},
\end{equation}
and Dice overlap between the cartilage mask and the top 5\% or 10\% salient voxels (Dice@5/10). mass@ROI is the fraction of total saliency mass inside cartilage; top1@ROI is the proportion of the top 1\% salient voxels that fall inside the cartilage mask. These measures quantify how much attribution falls inside or overlaps the cartilage region and are treated as descriptive post-hoc evidence.

\subsection{Statistical comparison}

All predictive comparisons are paired on the same held-out subjects. OA accuracy is compared with McNemar's test (exact binomial when the number of discordant pairs is below 25, otherwise chi-square with continuity correction). AUC-, F1-, and KL-related metrics are compared using paired subject-level bootstrap estimates of the mean performance difference with $B=10{,}000$ resamples (fixed seed 42), percentile 95\% confidence intervals, and two-sided bootstrap $p$-values. Dual is compared with Single-OA for OA metrics and with Single-KL for KL metrics. This paired design isolates the effect of supervision setting under the same test cases. The 18 resulting Dual-versus-single comparisons (3 backbones $\times$ 6 metrics) form one pre-specified family; we report nominal $p$-values together with Benjamini--Hochberg adjusted $q$-values and use $q<0.05$ as the corrected significance threshold.

\section{Experiments}

\subsection{Predictive performance}

\begin{figure}[t]
    \centering
    \begin{subfigure}[t]{0.48\textwidth}
        \centering
        \includegraphics[width=\textwidth]{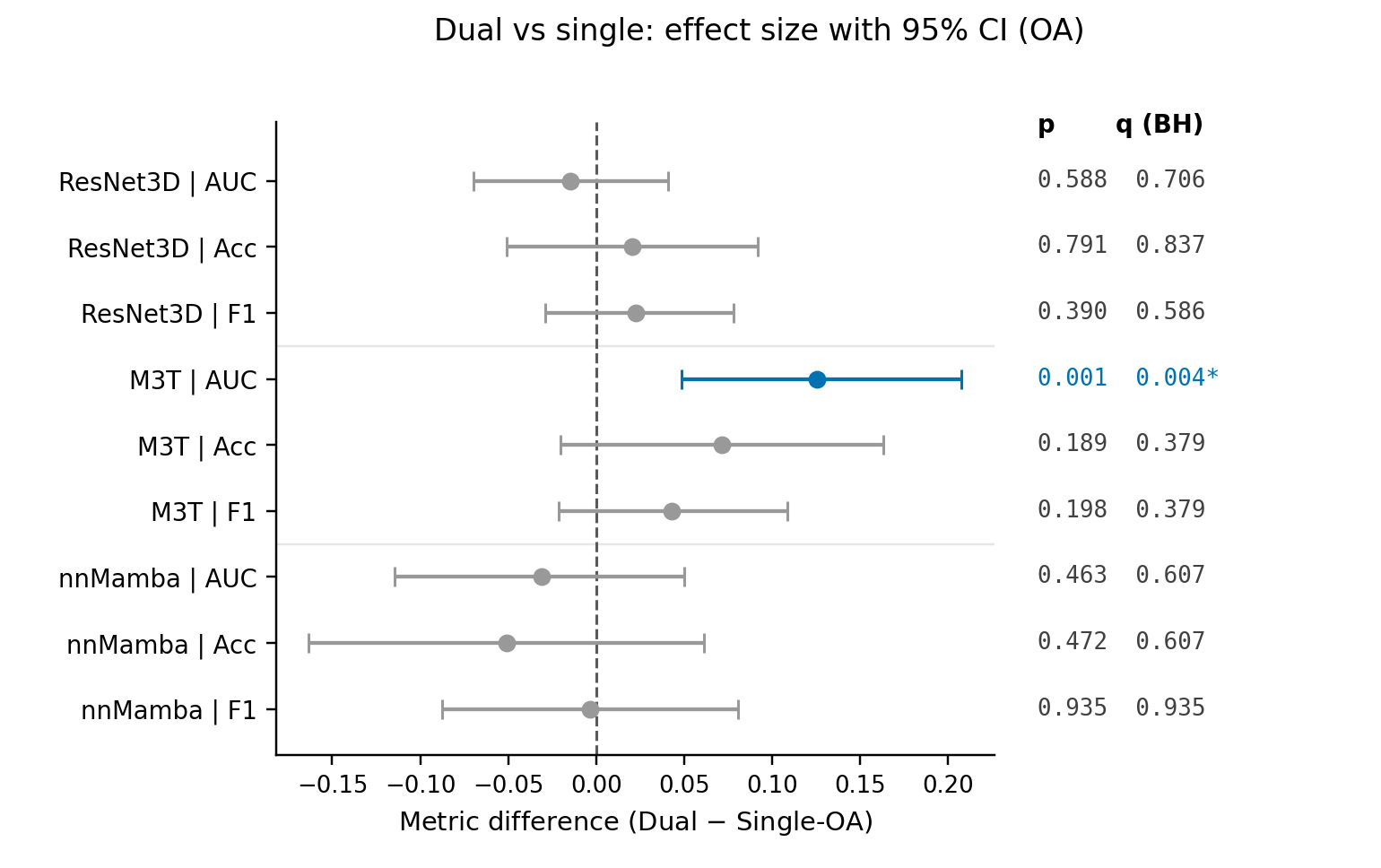}
        \caption{OA: Dual vs. Single-OA.}
    \end{subfigure}\hfill
    \begin{subfigure}[t]{0.48\textwidth}
        \centering
        \includegraphics[width=\textwidth]{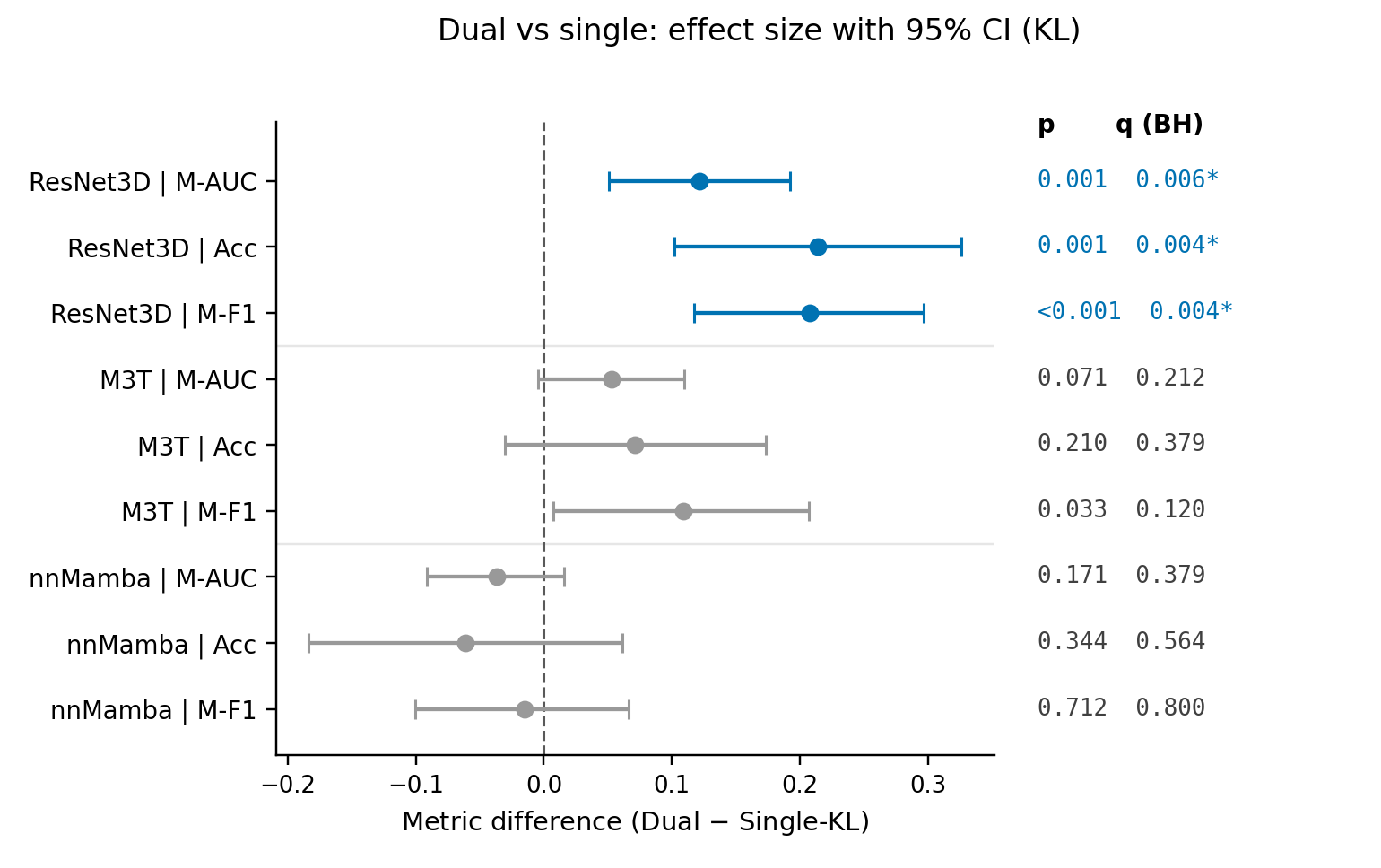}
        \caption{KL: Dual vs. Single-KL.}
    \end{subfigure}
    \caption{Paired effect sizes with 95\% confidence intervals. Positive values favor Dual. Nominal bootstrap $p$-values and Benjamini--Hochberg adjusted $q$-values are annotated; asterisks mark comparisons significant at $q<0.05$.}
    \label{fig:dual_single_effects}
\end{figure}

Table~\ref{tab:main_results} reports the main predictive results and Fig.~\ref{fig:dual_single_effects} shows paired effect sizes with 95\% confidence intervals. For ResNet3D, OA metrics remain statistically comparable to Single-OA. Dual substantially improves KL grading over Single-KL, with gains in macro-AUC ($\Delta=0.1219$, $p=0.001$, $q=0.006$), accuracy ($\Delta=0.2143$, $p<0.001$, $q=0.004$), and macro-F1 ($\Delta=0.2079$, $p<0.001$, $q=0.004$), all significant after FDR correction. M3T shows a narrower response: Dual improves OA AUC ($\Delta=0.1254$, $p<0.001$, $q=0.004$), while the nominal KL macro-F1 gain ($\Delta=0.1089$, $p=0.033$) does not survive FDR correction ($q=0.120$) and the KL macro-AUC gain is not nominally significant ($\Delta=0.0530$, $p=0.071$). nnMamba shows small or negative effects without significant improvement. These results establish that the effect of coarse-to-fine supervision depends on backbone architecture and task.

\begin{table}[t]
\centering
\small
\caption{Predictive performance across backbones and supervision settings.}
\label{tab:main_results}
\begin{tabular}{llcccccc}
\toprule
\multirow{2}{*}{Backbone} & \multirow{2}{*}{Setting} & \multicolumn{3}{c}{OA} & \multicolumn{3}{c}{KL} \\
\cmidrule(lr){3-5}\cmidrule(lr){6-8}
 & & AUC & Acc & F1 & M-AUC & Acc & M-F1 \\
\midrule
ResNet3D & Single-OA & \textbf{.881} & .796 & .851 & -- & -- & -- \\
 & Single-KL & .671 & .622 & .694 & .673 & .367 & .288 \\
 & Dual & .866 & \textbf{.816} & \textbf{.873} & \textbf{.794} & \textbf{.582} & \textbf{.496} \\
\midrule
M3T & Single-OA & .792 & .724 & .806 & -- & -- & -- \\
 & Single-KL & .818 & .765 & .832 & .723 & .388 & .358 \\
 & Dual & \textbf{.917} & \textbf{.796} & \textbf{.848} & \textbf{.776} & \textbf{.459} & \textbf{.467} \\
\midrule
nnMamba & Single-OA & .850 & .755 & .803 & -- & -- & -- \\
 & Single-KL & \textbf{.869} & \textbf{.765} & \textbf{.844} & \textbf{.749} & \textbf{.388} & \textbf{.246} \\
 & Dual & .819 & .704 & .800 & .713 & .327 & .231 \\
\bottomrule
\end{tabular}
\end{table}

\subsection{Representation geometry}

Table~\ref{tab:severity_axis_summary} summarizes the dominant latent severity axis, with PCA and the PC1 probe fitted on training subjects and evaluated on the held-out test set. For ResNet3D, Single-OA produces the strongest absolute PC1 alignment, consistent with a representation dominated by the coarse disease partition. Dual increases PC1--KL alignment relative to Single-KL from $0.666$ to $0.753$ and improves the PC1 probe AUROC from $0.843$ to $0.873$, indicating that the shared representation preserves a stronger fine-grained severity association than KL-only supervision. For M3T, Dual yields the strongest alignment with KL severity ($0.722$) and OA status ($0.679$), together with the highest OA probe AUROC ($0.915$). nnMamba shows the opposite pattern: Single-KL has stronger KL/OA alignment than Dual ($0.781/0.605$ vs. $0.689/0.522$), and its predictive metrics also decline under Dual. The geometry analysis therefore supports the same architecture-dependent pattern observed in the classification results. Importantly, correlation sign itself is not interpreted because PCA axis orientation is arbitrary.

\begin{table}[t]
\centering
\scriptsize
\caption{Severity-axis geometry. Standardization, PCA, and the PC1 probe are fitted on the 383 training subjects; all reported values are computed on the 98 held-out test subjects. Correlations are reported as magnitudes because PC1 sign is arbitrary.}
\label{tab:severity_axis_summary}
\begin{tabular}{llcccc}
\toprule
Backbone & Setting & EVR$_{\mathrm{PC1}}$ & $|\rho_{\mathrm{KL}}|$ & $|\rho_{\mathrm{OA}}|$ & AUROC$_{\mathrm{OA}}$ \\
\midrule
\multirow{3}{*}{ResNet3D}
& Single-OA & \textbf{.978} & \textbf{.802} & \textbf{.628} & \textbf{.883} \\
& Single-KL & .301 & .666 & .562 & .843 \\
& Dual & .423 & .753 & .610 & .873 \\
\midrule
\multirow{3}{*}{M3T}
& Single-OA & \textbf{.996} & .647 & .514 & .814 \\
& Single-KL & .546 & .683 & .543 & .832 \\
& Dual & .606 & \textbf{.722} & \textbf{.679} & \textbf{.915} \\
\midrule
\multirow{3}{*}{nnMamba}
& Single-OA & .936 & .628 & .574 & .851 \\
& Single-KL & \textbf{.970} & \textbf{.781} & \textbf{.605} & \textbf{.869} \\
& Dual & .967 & .689 & .522 & .819 \\
\bottomrule
\end{tabular}
\end{table}

\subsection{Saliency--cartilage overlap}

\begin{figure}[t]
    \centering
    \includegraphics[width=0.86\textwidth]{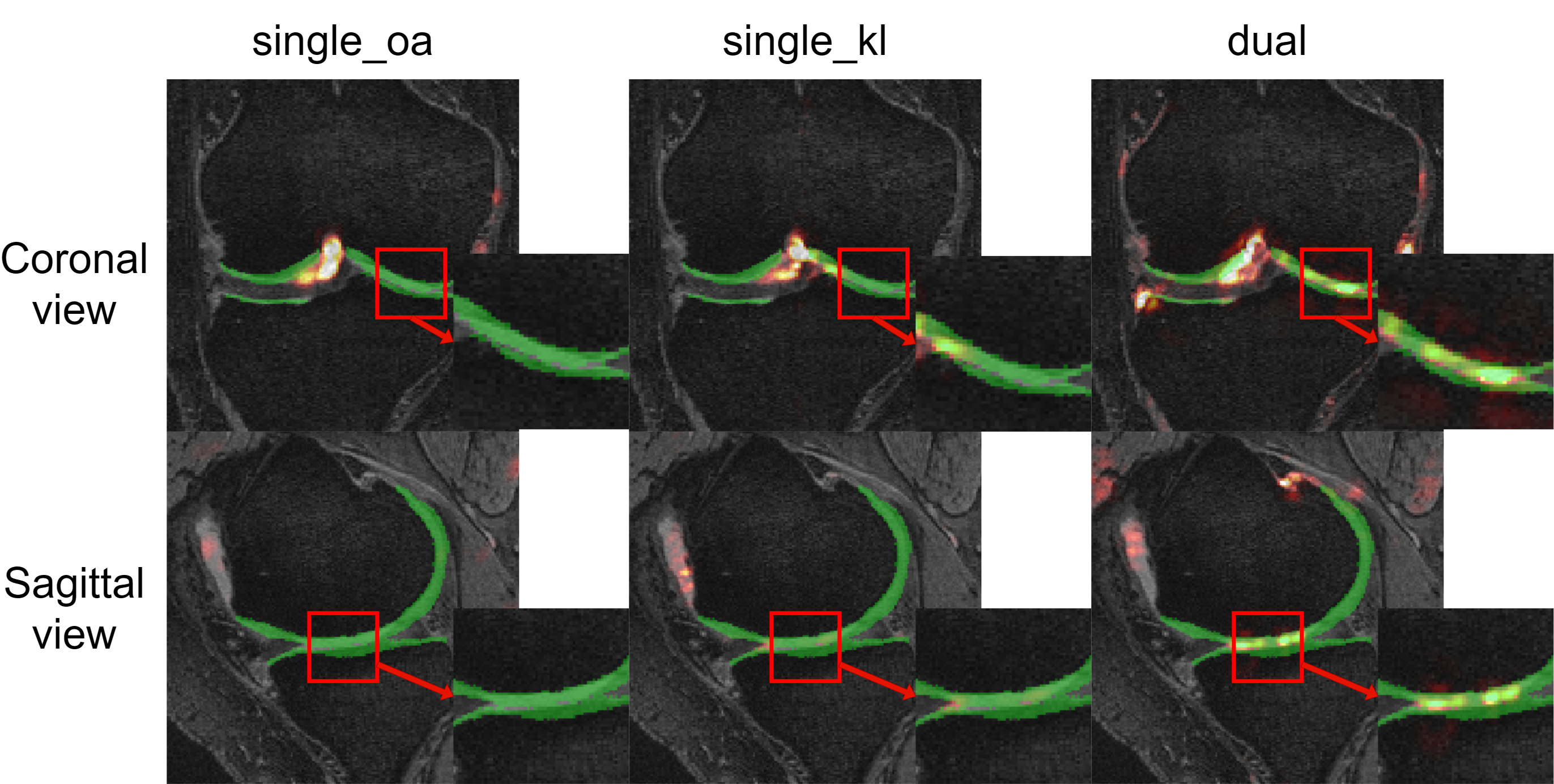}
    \caption{Representative saliency maps overlaid with cartilage regions for single-task and Dual settings.}
    \label{fig:saliency_cartilage}
\end{figure}

Table~\ref{tab:saliency_overlap} provides descriptive attribution results. ResNet3D shows the clearest change: both Dual branches increase cartilage overlap relative to their single-task counterparts, and Dual-OA reaches the highest values across mass@ROI, top1@ROI, Dice@5, and Dice@10. M3T exhibits a smaller but consistent increase in several overlap measures. nnMamba again differs from the responsive backbones, with both Dual heads decreasing on most overlap metrics. Figure~\ref{fig:saliency_cartilage} illustrates representative spatial patterns and highlights the stronger concentration near cartilage for the responsive backbones. The analysis uses input-gradient attribution without a matched-random or mask-size-matched baseline, so these values are treated as descriptive evidence of attribution patterns with claims limited to spatial association.

\begin{table}[t]
\centering
\scriptsize
\caption{Descriptive saliency--cartilage overlap.}
\label{tab:saliency_overlap}
\begin{tabular}{llcccc}
\toprule
Backbone & Setting & mass@ROI & top1@ROI & Dice@5 & Dice@10 \\
\midrule
\multirow{4}{*}{ResNet3D}
& Single-OA & .0793 & .1769 & .1608 & .1123 \\
& Single-KL & .0475 & .1209 & .1293 & .0998 \\
& Dual-KL & .0861 & .2243 & .1679 & .1150 \\
& Dual-OA & \textbf{.0941} & \textbf{.2320} & \textbf{.1775} & \textbf{.1199} \\
\midrule
\multirow{4}{*}{M3T}
& Single-OA & .0691 & .0817 & .0833 & .0693 \\
& Single-KL & .0696 & .0877 & .0885 & .0695 \\
& Dual-KL & .0720 & \textbf{.1043} & \textbf{.0938} & \textbf{.0715} \\
& Dual-OA & \textbf{.0722} & .1036 & .0930 & .0713 \\
\midrule
\multirow{4}{*}{nnMamba}
& Single-OA & .0754 & \textbf{.1904} & \textbf{.1666} & .1155 \\
& Single-KL & .0713 & .1811 & .1635 & \textbf{.1161} \\
& Dual-KL & .0564 & .1015 & .1195 & .0972 \\
& Dual-OA & .0568 & .1025 & .1199 & .0973 \\
\bottomrule
\end{tabular}
\end{table}

\section{Discussion}

The experiments reveal a consistent architecture-dependent pattern. ResNet3D shows the clearest benefit from Dual supervision: its KL-grading gains are the largest observed and remain significant after FDR correction, accompanied by stronger fine-grained severity alignment than Single-KL. M3T shows a narrower response, with a significant OA AUC gain and nominal KL improvements that do not survive FDR correction. nnMamba behaves differently: its Single-KL representation already exhibits strong severity alignment and its Dual model does not improve predictive performance. This negative result is informative because it shows that the coarse objective is not universally beneficial and that its effect depends on how the backbone shares and organizes task information.

The derived nature of the OA label helps clarify the interpretation of these results. The OA objective adds no new annotation information. Its contribution is an explicit coarse-granularity optimization constraint that groups KL 0--1 and KL 2--4. For responsive backbones, this additional objective appears to reinforce the clinically meaningful OA boundary while preserving enough within-branch variation for KL grading. The observed architecture dependence suggests that the interaction between coarse and fine objectives is mediated by backbone-specific optimization and feature-sharing dynamics. Dedicated analyses of gradient alignment, task interference, and training trajectories could test these mechanisms directly.

The representation results also motivate caution in how low-dimensional geometry is interpreted. PC1 provides a compact description of the dominant latent axis, while its sign has no intrinsic meaning. We therefore use correlation magnitude to quantify alignment strength and interpret PCA together with predictive performance. The saliency analysis provides a complementary anatomical view: ResNet3D and M3T show higher cartilage overlap under Dual, while nnMamba shows the opposite pattern. These overlap values are descriptive because gradient-based attribution can be unstable, cartilage represents only one OA-relevant structure, and the current analysis does not include a matched-random attribution baseline.

Several limitations define the scope of the conclusions. The experiments use one 3D knee MRI cohort with a predefined held-out split, supporting controlled within-cohort comparisons of supervision strategies while leaving cross-dataset generalizability for future work. Each configuration was trained once without explicit random-seed control, so run-to-run training variance is not characterized; the paired statistical design controls for test-case composition but not for training stochasticity. The KL head uses standard cross-entropy and therefore does not explicitly model ordinal distances between grades. PCA captures only selected aspects of high-dimensional representation structure. Future work can extend the current analysis with ordinal objectives, repeated or external validation, attribution robustness checks, and direct investigation of task-gradient interactions.

\section{Conclusion}

This study examines coarse-to-fine OA/KL supervision as a representation-learning problem in 3D knee MRI. Across three backbone families, Dual supervision improves prediction and severity-related latent organization for selected architectures, with the strongest gains observed for ResNet3D and more moderate gains for M3T. nnMamba provides a contrasting case in which single-task learning remains stronger. Together, predictive, latent-geometry, and cartilage-attribution analyses show that the effect of hierarchical supervision is measurable beyond scalar accuracy and is strongly conditioned by backbone architecture. These findings motivate supervision granularity as a useful design dimension for learning disease representations from noisy hierarchical clinical labels.



\bibliographystyle{splncs04}
\bibliography{references}

\end{document}